%% file: iclr2027_conference.tex
\PassOptionsToPackage{table}{xcolor}  
\documentclass{article} 
\usepackage{iclr2027_conference,times}
\usepackage{xcolor}  
\definecolor{medprimary}{HTML}{1F5F7A}
\definecolor{medtint}{HTML}{E8EFF2}

\usepackage[utf8]{inputenc} 
\usepackage[T1]{fontenc}    
\usepackage{hyperref}       
\usepackage{url}            
\usepackage{booktabs}       
\usepackage{amsfonts}       
\usepackage{nicefrac}       
\usepackage{microtype}      

\usepackage{multirow}   
\usepackage{makecell}   
\usepackage{graphicx}
\usepackage{svg}
\usepackage{amsmath}
\usepackage{array}  
\usepackage{algorithm}
\usepackage{algorithmicx} 
\usepackage{algpseudocode}
\usepackage{enumitem}
\usepackage{amssymb} 
\usepackage{tikz}
\usetikzlibrary{positioning, arrows.meta, fit, backgrounds, calc}
\usepackage{microtype}
\usepackage{siunitx}

\usepackage{calc}
\newlength{\passw}
\newcommand{\medaka}{\textsc{Medaka}}
\newcommand{\liquidworld}{\textsc{LiquidWorld}}

\usepackage{listings}
\usepackage{caption}
\usepackage{tcolorbox}
\tcbuselibrary{breakable, listings}

\title{Means-Ends Analysis Helps LLMs Plan in Continuous-State Domains}

\author{Jiayi Li, Carl Kingsford\\
Ray and Stephanie Lane Computational Biology Department\\
Kenneth C. Griffin School of Computer Science\\
Carnegie Mellon University\\
Pittsburgh, PA 15213, USA \\
\texttt{\{lijiayi, carlk\}@cs.cmu.edu} 
}

\iclrfinalcopy 
\begin{document}

\maketitle

\begin{abstract}
LLM planning research has traditionally focused on discrete-state domains, where actions form a finite, enumerable set. Existing LLM planning frameworks largely assume actions are enumerable, modeling planning as a search process over candidate actions. However, in continuous-state domains, where actions form a continuum and cannot be enumerated, this assumption breaks down. We find a remedy in Newell and Simon's Means-Ends Analysis (MEA), a general problem-solving technique rooted in heuristics from human psychology. We develop \medaka, an MEA-inspired structured prompting framework that encourages recursive goal decomposition, algebraic derivation, and per-step state tracking. To evaluate native LLM planning abilities in continuous-state domains, we develop \liquidworld, a systematic testbed on liquid mixing. On \liquidworld, \medaka\ can improve performance across model families, suggesting that classical means-ends analysis elicits LLM-native reasoning abilities that are helpful for continuous-state planning.

\end{abstract}

\input{sections/introduction}
\input{sections/related_work}
\input{sections/method}
\input{sections/experiments}
\input{sections/discussion}
\section*{AI use statement}

In this work, we used generative AI tools to generate synthetic data sets.
Specifically, the ``linear''-tagged problems' generation and verification pipeline is executed by an AI coding agent, using hand-implemented code.
We have not used generative AI tools to propose or refine hypotheses, design or provide feedback on research methodology or experiments, implement methods, assist with translation, clean or reformat dataset, support qualitative and thematic data analysis, interpret results,
and the rest of the required disclosure tasks (help develop theoretical models or conceptual frameworks, formulate mathematical claims, provide critical ingredients for proving mathematical claims, assist in the writing of proofs,) are not applicable to this work.

Additionally, we used generative AI tools to edit a research paper to improve readability. We have reviewed all AI-assisted work. LLM-generated synthetic data have all been verified for solvability by the authors. We take responsibility for the final content of this work, including text, claims or artifacts produced with the aid of generative AI.

\section*{Ethics statement}
The authors of this paper adhere to the ICLR Code of Ethics in every way.

\section*{Reproducibility statement}
As efforts made to ensure reproducibility, we presented the prompts for the methods as well as LLM configurations in the Appendices, and open-sourced the code with a link present at the introduction section.

\bibliographystyle{iclr2027_conference}
\bibliography{references}

\clearpage
\appendix
\input{sections/app_all_other}
\input{sections/app_prompts}

\end{document}

%% file: sections/introduction.tex
\section{Introduction}
Large language models (LLMs) have shown remarkable abilities across diverse domains, including commonsense reasoning~\citep{zellers2019hellaswag}, mathematics~\citep{wei2022chain}, and code generation~\citep{chen2021evaluating}. These successes have prompted interest in applying LLMs to more complex cognitive tasks, such as planning~\citep{valmeekam2023planbench}. 

Existing methods for LLM planning are primarily developed for tasks with discrete states and enumerable actions~\citep{valmeekam2023planbench}. 
These methods mainly build on an enumerate-and-rank assumption, where they model planning as searching actions over a discrete, enumerable space, and use techniques like state-aware backtracking~\citep{sel2025llms} or external tree structures~\citep{yao2023tree, hao2023reasoning} to aid the search process. 
However, in many scenarios such as biochemical experimentation, states are continuous, and actions are no longer enumerable. 
Such a distinction requires LLM capabilities beyond search: while actions cannot be enumerated, they can often be derived algebraically, given a faithful representation of the current state and its deviation from the goal state. 
In this case, planning natively in such domains requires an LLM to (1) decompose goals into tractable subgoals, (2) faithfully track states across steps, and (3) perform accurate derivations at each step.

An answer to this challenge comes from classical Means-Ends Analysis (MEA)~\citep{newell1963gps}, a general problem-solving technique rooted in human psychology. 
MEA is structurally well-suited to continuous action spaces because it treats undetermined parameters not as candidates to be enumerated and ranked, but as gaps to be closed by reasoning about the constraints relevant to the current subgoal.
Concretely, at each step, MEA (1) identifies a gap between the current state and subgoal, (2) proposes an operator (i.e., an action with potentially undetermined parameters) that would close it, (3) recursively resolves any precondition or parameter the operator leaves undetermined, treating each as a new subgoal, and (4) terminates when every parameter is determined. 
Because the gap is evaluated at each step, MEA also encourages state tracking: the state must be updated and re-examined after every operator application before the next subgoal is identified. 
In MEA, an undetermined parameter is a gap to be closed by accurately deriving a small system of equations linking the operator's parameters to a local subgoal, rather than requiring the LLM to reason about all constraints simultaneously.

Inspired by MEA, we develop \medaka, a structured prompting framework for LLM planning in continuous-state domains that encourages recursive goal decomposition and per-step state tracking. 
To isolate and evaluate LLM native planning abilities in continuous-state domains, we develop \liquidworld, a systematic testbed grounded in fluid mixing, a primitive operation to which numerous standard laboratory protocols (e.g., serial dilution, growth medium preparation, and buffer preparation) can be reduced. 
\liquidworld\ problems cover diverse structural characteristics, isolating key reasoning capabilities for planning in continuous-state domains (Table~\ref{tab:levels}): nonlinearity of the mixing constraints, coupling of action parameters, and context length (number of goals and distractor vessels).
On \liquidworld, \medaka\ can yield performance improvement over existing prompting frameworks for LLM planning across both closed-source (Claude~\citep{anthropic2026sonnet46}, GPT~\citep{hurst2024gpt, singh2025openai}, Gemini~\citep{google2025gemini3flash}) and open-source model families (Llama~\citep{grattafiori2024llama}, Gemma~\citep{team2026gemma}), suggesting that the enumerate-and-rank assumption shared by existing methods is mismatched with continuous-state planning, and that classical means-ends analysis offers a better structural fit. We make our code available at \url{https://anonymous.4open.science/r/medaka_code-FFD0}.



Our contributions are as follows:
\begin{itemize}[]
\item We identify continuous-state planning as a distinct challenge for LLM planning and characterize the enumerate-and-rank assumption as an architectural mismatch with this setting.
\item We introduce \liquidworld, a systematic testbed for isolating and evaluating LLM native planning in continuous domains, covering diverse problem structures in fluid mixing.
\item We propose \medaka, a structured prompting framework for LLM planning grounded in Means-Ends Analysis that encourages recursive goal decomposition and per-step state tracking, that can better elicit LLM native planning abilities on \liquidworld.
\end{itemize}

%% file: sections/related_work.tex
\section{Related work}
\paragraph{LLM planning frameworks.}
Current LLM planning frameworks mainly elicit LLM native planning behavior via prompting. 
Chain-of-Thought (CoT)~\citep{wei2022chain} prompting instructs the model to articulate intermediate reasoning steps before committing to an action and is the default planning framework for modern LLMs. 
Tree-of-Thoughts (ToT)~\citep{yao2023tree} extends CoT with explicit branching and value-guided search over partial solutions. 
RAP~\citep{hao2023reasoning} casts the LLM as a world model and runs Monte Carlo Tree Search over its predicted transitions. 
Algorithm of Thoughts (AoT)~\citep{sel2024aot} embeds a simulated search trace inside the prompt, while AoT+~\citep{sel2025llms} augments this with periodic state memoization to mitigate state hallucination. 
Another line of research in LLM-Modulo~\citep{kambhampati2024llmmodulo} frameworks equips LLMs with external solvers or verifiers to offload parameter derivation. 
For example, LLM3~\citep{wang2024llm3} uses a motion planner to validate LLM-proposed parameterized actions, and PRoC3S~\citep{curtis2024proc3s} uses the LLM to output a constraint-satisfaction program whose parameters are filled in by a constraint-satisfaction solver. 
The goal of our work is to isolate and elicit native LLM planning abilities in continuous-state domains, not to build a deployable planner. 
Hence, we do not compare with LLM-Modulo methods, since their external solvers perform the parameter derivations that we aim to measure natively in LLMs.

\paragraph{LLMs in the laboratory.}
A growing body of research employs LLMs as agents in scientific laboratories. 
Coscientist~\citep{boiko2023autonomous} equips GPT-4 with tools in internet search, code execution, and robotic instrumentation to autonomously design and run chemistry experiments.
ChemCrow~\citep{m2024augmenting} augments GPT-4 with 18 expert-designed tools spanning organic synthesis, drug discovery, and materials design, demonstrating the ability to automate the synthesis of small molecules. Reinforcement learning has also been applied to protocol planning for robotic liquid handlers, a problem that is NP-complete~\citep{ferdosi2023reinforcement}.
These systems show that continuous quantities arise naturally when LLMs operate in laboratory settings. 
They rely on external tools, agentic orchestration, or task-specific training, whereas we study the planning ability of the LLM itself.

\paragraph{Classical planning and Means-Ends Analysis.}
Classical planning provides the formal vocabulary (e.g., operators, preconditions, decomposition) that our framework adapts to LLM-based reasoning.
The General Problem Solver~\citep{newell1963gps} introduces Means-Ends Analysis as a recursive difference-reduction procedure. 
STRIPS~\citep{fikes1971strips} formalized symbolic operators with preconditions and effects, 
and PDDL~\citep{mcdermott1998pddl} standardized this representation as a planning language. 
PDDL 2.1~\citep{fox2003pddl21} extended the language with numeric fluents, 
and Metric-FF~\citep{hoffmann2003metricff} provided a heuristic solver for the resulting numeric planning problems. 
HTN~\citep{erol1994htn} planning introduced hierarchical task networks that recursively decompose compound tasks into primitives. 
Together, these works supply the symbolic substrate that \medaka\ imports into LLM prompting.


%% file: sections/method.tex
\section{Methods}
\label{sec:method}

\subsection{Problem formulation}
\label{sec:problem}

\paragraph{From discrete to continuous-state planning.}
LLM planning has been studied predominantly in \emph{discrete-state} domains such as BlocksWorld~\citep{valmeekam2023planbench}, where state transitions are symbolic, not algebraic. 
LLM planning frameworks, e.g., Chain-of-Thought~\citep{wei2022chain}, Tree-of-Thoughts~\citep{yao2023tree}, RAP~\citep{hao2023reasoning} and AoT$+$~\citep{sel2025llms} have advanced the state of the art on this classic benchmark, but their core mechanisms (tree search, beam search, in-context simulated search) presuppose a finite action space. 
However, many planning domains in biochemical laboratories, e.g., liquid handling, involve states governed by continuous numerical quantities. 
Whether LLM planning methods designed for discrete action selection transfer to such continuous-state domains has not been formally investigated.

\paragraph{Planning formulation.}
Let $\mathcal{S}$ be a state space, $\mathcal{A}$ an action space, and
$\mathcal{T}\colon \mathcal{S}\times\mathcal{A}\rightharpoonup\mathcal{S}$
a transition function.
A planning problem is a tuple
$\mathcal{P}=(\mathcal{S},\mathcal{A},\mathcal{T},s_0,g)$, where
$s_0\in\mathcal{S}$ is the initial state and
$g\colon\mathcal{S}\to\{0,1\}$ is a goal
predicate that identifies goal states~\citep{ghallab2004automated}.
We denote $\mathcal{A}(s)=\{a\in\mathcal{A}:\mathcal{T}(s,a)\text{ is defined}\}$
as the set of actions applicable in~$s$.
A \emph{plan} is a sequence $(a_1,\dots,a_T)$ such that, setting
$s_t=\mathcal{T}(s_{t-1},a_t)$ for each~$t$, every action is applicable
($a_t\in\mathcal{A}(s_{t-1})$) and the final state satisfies the goal
($g(s_T)=1$).
When the planner is an LLM~$\pi$, each generated step implicitly performs
two sub-tasks: 

\begin{itemize}
  \item \textbf{Action selection}: choosing
    $a_t\!\in\!\mathcal{A}(s_{t-1})$ to reduce the gap between $s_{t-1}$
    and~$g$;
  \item \textbf{State prediction}: computing
    $\hat{s}_t\approx\mathcal{T}(s_{t-1},a_t)$, i.e.\ simulating the
    transition in natural language.
\end{itemize}

\noindent This decomposition is central to our analysis: existing methods
differ primarily in how they handle action selection, while leaving state
prediction to the LLM's unstructured generation.
In discrete domains $\mathcal{A}(s)$ is finite for every~$s$, and planning
reduces to a search over a finite graph.
In continuous-state domains,
$\mathcal{A}(s)$ is a continuum, thus enumerating candidate actions is no longer possible; a valid plan may
require parameter values that no finite discretisation of~$\mathcal{A}$
contains. The planner must
\emph{derive} the correct~$v$ from the current state and goal, not select
it from a finite list.
On the state-prediction side, once $a_t$ is fixed, computing
$\hat{s}_t = \mathcal{T}(s_{t-1}, a_t)$ becomes a numerical task that
existing LLM planning frameworks leave unstructured.

We study continuous-state LLM planning along two axes: 
\begin{itemize}
    \item \textbf{LiquidWorld} (\S\ref{sec:liquidworld}), a systematic testbed for LLM continuous-state planning, grounded in fluid mixing tasks with varying structural characteristics.
    \item \textbf{MEDAKA} (\S\ref{sec:med}), a structured prompting framework for LLM continuous-state planning, inspired by Means-Ends Analysis~\citep{newell1963gps}.
    \medaka\ recursively decomposes each goal until all numerical
    parameters are derived \emph{before} the action is committed.
\end{itemize}


\subsection{\liquidworld: A systematic testbed for continuous-state planning}
\label{sec:liquidworld}

To study continuous-state LLM planning under controlled conditions, we introduce
\liquidworld, a systematic testbed where the state space, action space,
and goal structure are held fixed across instances, while varying structural characteristics such as the number of goal states, nonlinearity of the liquid-mixing constraints, and the presence of distractors add to the difficulty of each instance (Table~\ref{tab:levels}). 
\liquidworld\ is built on fluid mixing, which is the foundation for numerous procedures in biochemical laboratories~\citep{ferdosi2023reinforcement} (e.g., serial dilution, growth medium preparation, and buffer preparation). 

\paragraph{Domain description.}
A \liquidworld\ state~$s$ consists of $n$~vessels $\{V_1,\dots,V_n\}$.
Each vessel~$V_i$ is characterized by a current volume
$\mathit{vol}_i\in\mathbb{R}_{\ge 0}$ and a concentration vector
$\mathbf{c}_i=(c_i^1,\dots,c_i^K)$ over $K$~constituents, with $\sum_{k}c_i^k=1\; (c_i^k\geq 0)$ whenever $\mathit{vol}_i>0$.
A goal~$g$ specifies target volumes and concentrations for one or more designated vessels.
A vessel has a \emph{pure} composition if $c_i^k=1$ for some~$k$, and a
\emph{mixture} otherwise.
A ``pouring'' action in the domain is denoted as $\textsc{Pour}(v,V_i,V_j)$, which transfers solutions of volume
$v$ from~$V_i$ to~$V_j$ ($0<v\le\mathit{vol}_i$).
Pouring obeys \emph{mass conservation}: all constituents in the source
transfer in proportion to their current concentrations, yielding the post-action state:
\begin{equation}\label{eq:pour}
  \mathit{vol}_i' = \mathit{vol}_i - v,\qquad
  \mathit{vol}_j' = \mathit{vol}_j + v,\qquad
  c_j^{k\prime} = \frac{\mathit{vol}_j \cdot c_j^k + v\cdot c_i^k}
                        {\mathit{vol}_j + v}
  \;\;\forall\,k.
\end{equation}
%


\paragraph{Problem specification.}
\liquidworld\ instances are stored in structured JSON format (per-vessel volume and composition dictionaries). 
An equivalent textual rendering is then presented to the LLM to generate plans. 
Figure~\ref{fig:examples} shows an example instance in structured text.

\begin{figure}[t]
\centering
\footnotesize
\noindent\fbox{\parbox{0.94\linewidth}{%
\textit{State:}\;
\texttt{V1 (capacity 999L): 10\,L L1;\;
  V2 (capacity 999L): 10\,L L2;\;
  V3 (capacity 999L): 10\,L L3;\;
  V4 (capacity 999L): empty}\\[3pt]
\textit{Goal:}\;
\texttt{V4: 5\,L, 40\% L1, 40\% L2, 20\% L3}
}}
\caption{A \liquidworld\ instance in the structured text
format presented to the LLM.
The grammar of this format is given in Figure~\ref{fig:grammar}.}
\label{fig:examples}
\end{figure}

To measure reasoning capabilities key to planning in continuous-state domains, we develop problems with five structural characteristics (Table~\ref{tab:levels}): (1) whether to allow direct pouring from source to target (``linear'') or (2) require sources to always route through intermediate pools (``nonlinear''); (3) whether the underlying algebraic system has heavy coupling across different sources (``coupling''); (4) whether there is a large number of targets (``multi-goal''); and (5) whether there are excess vessels that are neither targets nor necessary for use (``distractors''). 

Each \liquidworld\ instance exhibits one or more such characteristics, which function as difficulty tags: (1) While pouring actions can be read out iteratively, one liquid at a time, when there is no coupling between the sources, algebraically solving a linear system is required for ``coupling'' problems. (2) ``linear'' problems are solvable under a single linear program (LP), which we use only to construct instances with verifiable ground truth, while ``nonlinear'' ones are more mathematically and structurally complex, since the compositions of the intermediate pools are unknown, resulting in bilinear goal constraints (to and from the pool). ``nonlinear'' instances are derived from the pooling problem~\citep{haverly1978studies}, which is an NP-hard process optimization problem~\citep{alfaki2013strong,haugland2016complexity}. (3) Problems tagged with ``multi-goal'' and ``distractors'' have longer context length than those that are not.

For ``linear''-tagged problems, we apply a parametric generator, whose input includes the grammar of the problem specification language (Figure~\ref{fig:grammar}) and the parameters for the start and goal states (number of source vessels, target vessels, and liquids). ``nonlinear''-tagged problems also conform to the above grammar but are hand-designed to account for the constraints of intermediate pools.

\begin{figure}[t]
\centering
\footnotesize
\begin{tabular}{@{}r@{\;\;}c@{\;\;}l@{}}
\toprule
\texttt{<state>}   & $::=$ & \texttt{<vessel\_decl> (";" <vessel\_decl>)*} \\
\texttt{<vessel\_decl>} & $::=$ & \texttt{NAME "(" "capacity" FLOAT "L" ")" ":" <contents>} \\
\texttt{<contents>} & $::=$ & \texttt{"empty"} \\
                     & $\mid$ & \texttt{FLOAT "L" LID}  \hfill\textit{// pure vessel}\\
                     & $\mid$ & \texttt{FLOAT "L" "," <pct> ("," <pct>)*}  \hfill\textit{// mixture}\\
\texttt{<pct>}      & $::=$ & \texttt{FLOAT "\%" LID} \\[4pt]
\texttt{<goal>}     & $::=$ & \texttt{<goal\_clause> (";" <goal\_clause>)*} \\
\texttt{<goal\_clause>} & $::=$ & \texttt{NAME ":" FLOAT "L" "," <pct> ("," <pct>)*} \\[4pt]
\texttt{<action>}   & $::=$ & \texttt{"Pour" VOL "L" "from" NAME "to" NAME} \\
\texttt{VOL}        & $::=$ & \texttt{FLOAT $\mid$ INT "/" INT} \\
\bottomrule
\end{tabular}
\caption{Grammar of the \liquidworld\ problem specification language.
\texttt{NAME} is a vessel identifier (e.g.\ \texttt{V1}),
\texttt{LID} a liquid identifier (e.g.\ \texttt{L2}),
and \texttt{FLOAT}/\texttt{INT} are numeric literals.
The grammar is implicit in the regex-based parser; we present it here
in Backus-Naur Form for clarity.}
\label{fig:grammar}
\end{figure}

\begin{table}[t]
\centering
\caption{\liquidworld\ structure. Each problem is annotated with one or more structurally defined difficulty tags.}
\label{tab:levels}
\renewcommand{\arraystretch}{0.95}
\footnotesize
\begin{tabular}{@{}lp{0.72\linewidth}r@{}}
\toprule
Tag & Criterion & \#Probs \\
\midrule
linear      & The volumes to pour from each source into each target vessel to reach the goal can be obtained by solving a linear system. & 136 \\
nonlinear   & Sources reach targets only through intermediate pools whose compositions are chosen by the plan; the pour volumes and intermediate compositions cannot be obtained from a linear system. & 53 \\
coupling  & Liquids recur across sources ($\#\textit{source--liquid pairs} - \#\textit{liquids} \ge 3$), so the pour volumes must be solved as one coupled system. & 106 \\
multi-goal  & $\ge 5$ target vessels. & 76 \\
distractor  & $\ge 2$ vessels that are neither targets nor used by the reference plan. & 28 \\
\midrule
Total       & & 189 \\
\bottomrule
\end{tabular}
\end{table}

\paragraph{Solvability verification.}
After problems are generated, we verify solvability in three stages before any instance enters the
\liquidworld\ testbed:
(1) A \emph{volume check} ensures that the total available volume of each liquid across all sources meets the total demand across all goals.
(2) A \emph{concentration check} ensures that no target requires a liquid at a concentration exceeding its maximum concentration across all sources.
(3) A \emph{linear-program (LP) feasibility check} (for ``linear'' problems only) confirms that a non-negative sequence of $\textsc{Pour}$ actions exists and serves as verifiable ground truth: the decision variables are the volumes poured from each source (to each target in the multi-target case), with equality constraints requiring each goal composition to be met exactly and inequality constraints capping the total volume drawn from each source. For multi-target problems, a joint LP is solved so that targets sharing sources are simultaneously satisfiable. Instances that fail any check are discarded and resampled. 
The Full LP formulations are given in Appendix~\ref{app:generator}.``nonlinear'' problems are hand-solved to check feasibility, after passing the previous two checks.


\paragraph{Plan validator.}
Requiring no external solvers or LLM-based judges, the validator parses a candidate plan into a sequence of $\textsc{Pour}$ actions, applies Eq.~\eqref{eq:pour} step by step, and checks whether each goal vessel matches its target.
Specifically, for every goal vessel, (i)~the absolute volume of each liquid component must match its target within $\epsilon = 0.05$\,L, and (ii)~the total vessel volume must be within the same tolerance. Target volumes range from $1$ to $5$\,L, so this corresponds to $1$--$5\%$ relative error. For "nonlinear" instances, it additionally checks that the liquid passed through the designated pool.

\subsection{\medaka: Means-Ends Decomposition for Continuous-State Planning}
\label{sec:med}
 
\begin{figure}[t]
\centering
\includegraphics[width=\textwidth]{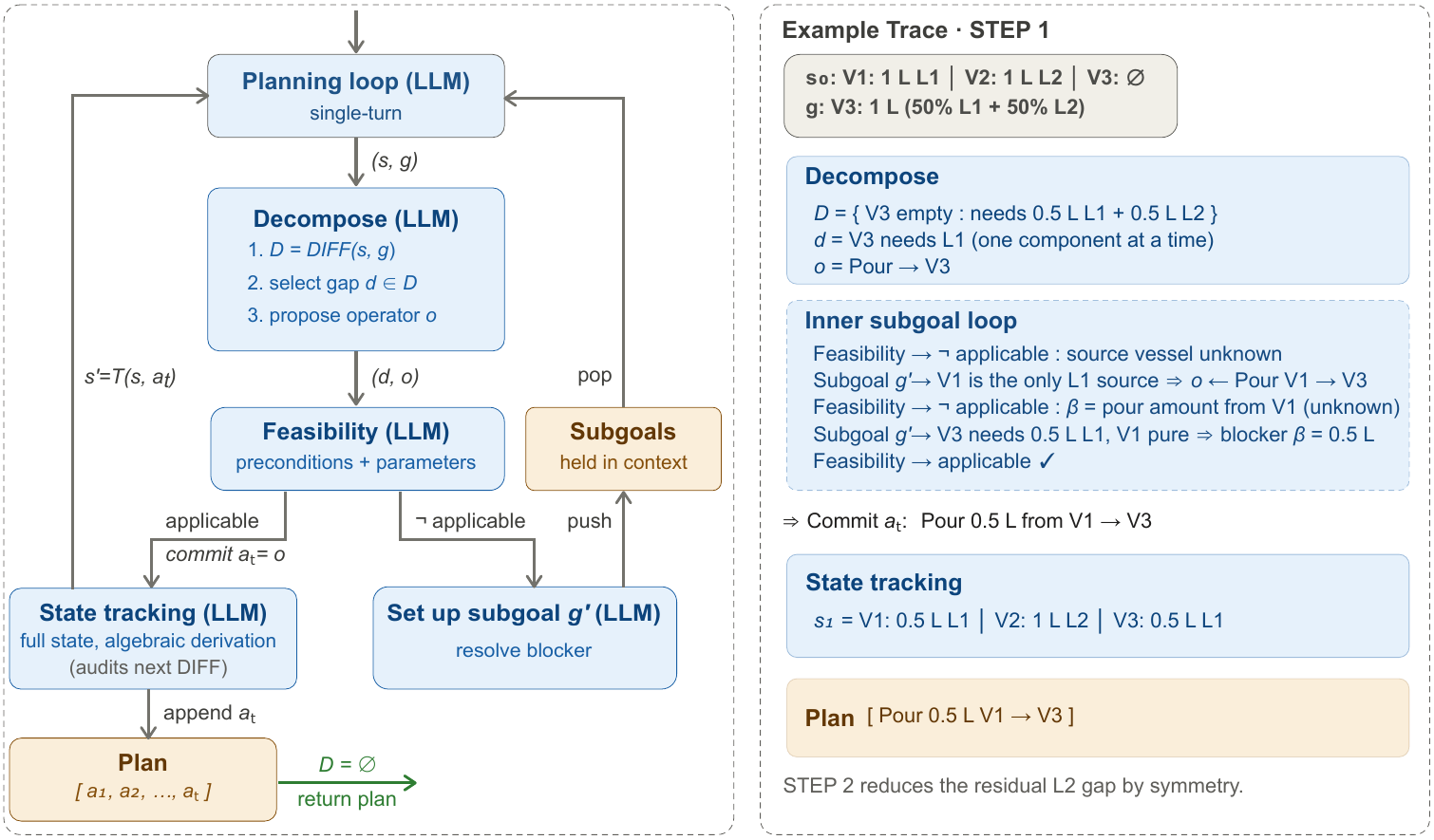}
\caption{Overview of \medaka. \textbf{Left:} the LLM recursively decomposes the goal by identifying a gap $d$ between the state $s$ and goal $g$ and proposes an operator $o$. A plan is returned when no gap remains. 
Feasibility tests whether $o$ is applicable in $s$: when blocked, the LLM derives a subgoal $g'$ that resolves the blocker algebraically and updates $o$, with deferred subgoals pushed onto a subgoal stack. 
Once applicable, an action $a_t = o$ is committed and the state tracker emits the full successor $s' = T(s, a_t)$.
\textbf{Right:} An example trace through STEP 1 of a \liquidworld\ problem.}
\label{fig:med-arch}
\end{figure}

Section~\ref{sec:problem} argued that continuous-state planning requires deriving action parameters rather than selecting actions from an enumerable set. 
To facilitate planning in the continuous-state domain, we adopt a primitive from Means-Ends Analysis (MEA)~\citep{newell1963gps}, a general problem-solving approach rooted in human psychology that is built around parameter derivation instead of search. 

MEA identifies the gap between the current state and the goal and proposes an operator that closes that gap. 
Here, an ``operator'' is an action but potentially with underspecified parameters. 
For example, an operator in \liquidworld\ can be ``Pour to V3'' but with the exact amount unknown.
If the operator's preconditions are not satisfied, or parameters are not yet derived, a recursive subgoal is created to meet these preconditions or derive these parameters. 
While the original MEA formulation was symbolic, its logic naturally extends to numerical settings: a missing real-valued action parameter is a gap to close, and is algebraically derivable from the current state and goal.

We instantiate MEA as a structured prompting protocol called \medaka\ (Figure~\ref{fig:med-arch}). \medaka\ realizes MEA through two mechanisms: \emph{recursive goal decomposition} drives macro control flow, and \emph{per-step state tracking} audits the traces to facilitate faithful derivations. 
The complete algorithmic logic of \medaka\ is presented in Algorithm~\ref{alg:med} and Appendix~\ref{app:prompt}.

\paragraph{Recursive goal decomposition.}
At each planning step, \medaka\ instructs the LLM to (1) compute $D = \textsc{Diff}(s_{t-1}, g)$, the set of unsatisfied goal conditions, (2) propose an operator $o$ that would reduce one of these gaps, and (3) test whether $o$ is \emph{applicable}, i.e., all of $o$'s preconditions satisfied and all of its numerical parameters determined.
If $o$ is not applicable, then the LLM recursively sets up subgoals whose satisfaction makes $o$ applicable. 
Recursion terminates when a fully parameterized, precondition-satisfying operator is reached, and only then does the LLM commit to an action $a$.  

This scheme encourages every undetermined parameter to be made an explicit subgoal and fully specified before an action is committed, rather than guessed and incrementally amended.
It places reasoning effort before commitment, reducing the propagation of errors into the generated plan. 
This marks a deviation from the workflow established by planning frameworks like AoT~\citep{sel2024aot} and AoT+~\citep{sel2025llms}, which place emphasis on backtracking when the plan goes awry.

\paragraph{Per-step state tracking.}
Per-step state tracking governs all numerical computation in \medaka, encouraging faithful algebraic derivations at every planning step.
It instructs the LLM to evaluate the transition equations (Eq.~\ref{eq:pour}) against the tracked state in both directions (Appendix~\ref{app:prompt}): 
backward, by setting up and solving equations for action parameters; forward, by deriving the complete successor state $s_t=\mathcal{T}(s_{t-1}, a_t)$ after every committed action $a_t$, listing the volume and composition of every vessel, not only those affected by the action.
This contrasts with AoT+, which caches states only at heuristically chosen intervals during its in-context simulated search.
In \medaka, the state output is instead a \emph{computational} mechanism that constrains subsequent reasoning: because the next step's \textsc{Diff} is computed from the recorded $s_t$, a hallucinated state either produces an inconsistent gap or surfaces as a contradiction in later steps.

%% file: sections/experiments.tex
\section{Experiments}
\subsection{Setup}
We compare \medaka\ against two prompting-based LLM planning methods: Chain-of-Thought (CoT)~\citep{wei2022chain} and AoT+~\citep{sel2025llms}. CoT is the standard sequential reasoning baseline that represents the default LLM reasoning behavior, while AoT+ is a recent state-of-the-art prompting framework for LLM planning that augments reasoning with in-context search and periodic state memoization. Both methods were originally developed and evaluated primarily in discrete symbolic planning domains.
We do not compare with LLM-Modulo methods since external solvers confound native parameter derivation abilities key to planning.
We evaluate across seven LLMs spanning both proprietary and open-weight families across different model sizes: \texttt{Claude-Sonnet-4.6, Gemini-3-Flash, GPT-5.4, GPT-4o, LLaMA-3.3-70B, Gemma4-4B}, and \texttt{Gemma4-2B}.
Since our question is which reasoning structure helps continuous-state planning, we disable thinking mode for all models so that their pretrained reasoning patterns do not override the tested prompting methods' intended behavior.
For each (model, method, problem) triplet, we perform three repeated rollouts and measure \texttt{pass@1} (accuracy) and \texttt{pass@3}, where a pass is determined by the plan validator described in Section~\ref{sec:liquidworld}.

\subsection{Main results}

\input{tables/main}    
As shown in Table~\ref{tab:accuracy_tags}, \medaka\ yields the best overall performance (under ties) on \liquidworld, highlighting its potential to enhance general LLM planning abilities in continuous-state domains. Our results also confirm that the structural characteristics induce problem difficulty: across all tested methods and models, problems tagged ``nonlinear'', ``coupling'', and ``multi-goal'' have pass rates substantially below the overall average. 

Problem structural difficulty tags show cross-model performance heterogeneity that reflects discrepancies in reasoning capabilities. For instance, while frontier models including \texttt{Claude-Sonnet-4.6, Gemini-3-Flash}, and \texttt{GPT-5.4} are saturated on ``linear'' problems, they exhibit significantly lower vanilla CoT performance on ``nonlinear'' problems, but applying \medaka\ substantially improves them. Notably, \texttt{Claude-Sonnet-4.6} reaches 87.4\% \texttt{pass@1}, suggesting it does possess the underlying reasoning capabilities needed to solve planning problems with nonlinear constraints, and \medaka\ successfully elicits these capabilities where CoT fails to.

In contrast, older and open-weights models including \texttt{GPT-4o, LLaMA-3.3-70B, Gemma4-4B}, and \texttt{Gemma4-2B} are less capable on ``linear'' problems, but \medaka\ successfully improves their performance while AoT+ cannot, though to a level that is still below the frontier models. 
However, for ``nonlinear'' problems, \medaka\ improves the closed-source \texttt{GPT-4o} (9.4\% to 15.7\%) but fails to yield meaningful improvement for the tested open-weights models, suggesting that they inherently lack the reasoning capabilities to solve these problems. 

For ``coupling'' problems, we observe across-the-board gains compared to vanilla CoT when applying \medaka, showing a larger improvement than AoT+ (except on \texttt{GPT-5.4}). Since ``coupling'' problems require algebraic solving of equation systems, such gains corroborate \medaka's design, where faithful algebraic derivation is encouraged.

Our results also show \medaka's potential in facilitating management and reasoning over long context. 
We observe substantial gains for ``multi-goal'' problems (\texttt{Claude-Sonnet-4.6}: 50.4\% to 89.5\%) except \texttt{GPT-4o} and \texttt{Gemma} models. 
\medaka\ also raises performance on problems with distractor vessels to a near-perfect 94.0\% for \texttt{Claude-Sonnet-4.6} and induces a near 4-fold improvement for \texttt{GPT-4o} (10.7\% to 42.9\%). 
For both difficulty tags, neither \medaka\ nor AoT+ improve \texttt{Gemma4-2B}, suggesting its long-context reasoning capabilities are limited by its size.

\begin{figure}[htp]
    \centering
    \includegraphics[width=\linewidth]{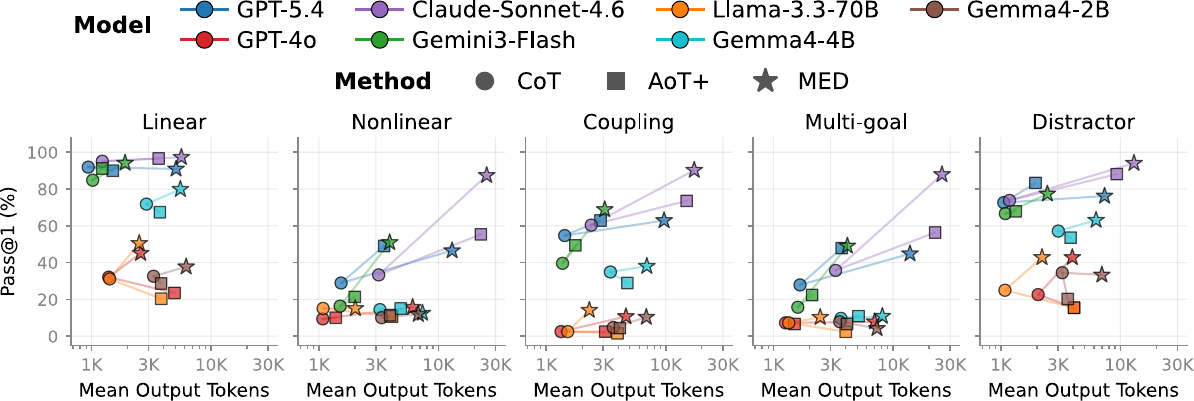}
    \caption{\texttt{pass@1} vs.\ mean output tokens on \liquidworld.
    Mean output tokens per trial (log-scale $x$) vs.\ Pass@1 ($y$). Markers denote methods, colors represent models, and thin lines connect methods of the same model. \medaka\ is abbreviated ``MED''.}
    \label{fig:token_pareto_output}
\end{figure}

Furthermore, Figure~\ref{fig:token_pareto_output} shows that \medaka\ uses more tokens than baselines while consistently improving performance (except \texttt{GPT-5.4}).
\medaka\ uses roughly 1.8 to 3.8$\times$ as many total tokens as CoT (median; Table~\ref{tab:tokens}), so we compare it against CoT with best-of-3 oracle selection (\texttt{pass@3}).
Even so, \medaka's \texttt{pass@1} exceeds CoT's \texttt{pass@3} on 4/7 models (all except \texttt{GPT-5.4} and the two \texttt{Gemma4} models); e.g., for \texttt{Claude-Sonnet-4.6} and \texttt{Gemini-3-Flash}, CoT reaches 86.8 and 67.2 \texttt{pass@3} (Table~\ref{tab:pass3_tags} in Appendix~\ref{app:pass3}), versus \medaka's 94.5 and 82.0 \texttt{pass@1}.
This suggests that, for most models, \medaka\ uses test-time compute more effectively than CoT.
   
In summary, these results suggest that \medaka\ can elicit LLM reasoning capabilities, e.g., resolving nonlinear constraints, faithful algebraic derivation, and long-context management/reasoning, that are crucial for planning in continuous-state domains.

\subsection{Qualitative failure modes}
To better understand how search-style prompts fail in continuous-state planning, we inspected failed trajectories on ``linear''-tagged problems from \texttt{GPT-4o} and \texttt{LLaMA-3.3-70B}, which are the two models where \medaka\ has the most improvement over AoT+ and CoT on these problems.
We analyze cases where CoT or AoT+ failed but \medaka\ succeeded or produced a substantially more consistent trace.
Table~\ref{tab:qualitative-failure-traces} in Appendix~\ref{app:failure} summarizes the dominant failure modes and provides example outputs.
The errors are not merely arithmetic slips: they reflect a mismatch between searching candidate actions and deriving continuous parameters, where valid pour volumes can only be determined by deriving mass-balance equations rather than enumeration.

\paragraph{Trial-and-error parameter search.}
A common AoT+ failure is to treat continuous pour volumes as quantities to be guessed, tested, and locally amended.
This behavior is encouraged by the backtracking structure of AoT+: when an intermediate state does not match the goal, the model often attempts another nearby action rather than re-deriving the entire set of pour parameters.
In the first example of Table~\ref{tab:qualitative-failure-traces}, the model reaches the target volume of $2$ L but with composition $(20,25,55)\%$ instead of $(13.43,30.52,56.05)\%$, then adds small corrective pours and attempts to undo the overshoot. 

\paragraph{State hallucination and drift.}
This appears in both CoT and AoT+ traces.
Although AoT+ explicitly summarizes intermediate states, this summary is an unstructured generation of the LLM and can become unreliable when state updates require numerical computation.
In the second example of Table~\ref{tab:qualitative-failure-traces}, the model empties $V_8$ but later reuses it as if it still contained liquid.

\paragraph{Locally correct but globally inconsistent arithmetic.}
A third failure mode is that the model performs locally correct computations while producing a globally inconsistent plan.
This occurs when the model reasons one action at a time rather than solving the coupled conservation constraints over all liquids simultaneously.
In the third example of Table~\ref{tab:qualitative-failure-traces}, the final volume reaches $4$ L, but the composition is $(36.7,40.0,23.3)\%$ instead of $(36.71,26.67,36.62)\%$.
The error arises because liquids can appear in multiple sources: increasing one component may also introduce excess of another component. 
Thus, satisfying the target requires solving the global coupled system, not greedily matching one component at a time.

\subsection{Ablation studies}\label{sec:ablation}
\input{tables/ablation}

We perform an ablation study on \medaka's two core components (recursive goal decomposition and per-step state tracking) using \texttt{Claude-Sonnet-4.6} and \texttt{Gemini-3-Flash}, the two models with the best \medaka\ performance and largest improvement over CoT.
We develop two ablated versions: ``$-$\,State Tracking'' drops the per-step state-tracking module, which includes all numerical computation instructions, and ``$-$\,Recursive Module'' drops all instructions related to recursive goal decomposition, retaining per-step tracking as an `algebra-aware' CoT. 
As shown in Table~\ref{tab:ablation}, removing either module reduces performance, suggesting that the recursive decomposition (even without any numerical computation instructions) and state tracking modules each help LLMs on continuous-state planning.
Note that since each ablation retains one module, their contributions need not be additive. 

%% file: tables/main.tex
\newcommand{\tbd}{\textcolor{gray}{--}}

\begin{table}[ht]
\centering
\caption{\texttt{pass@1}(\%) on \liquidworld\ grouped by difficulty tag, computed as the mean pass rate over all rollouts on problems with that tag, $3$ rollouts per problem. A problem may carry one or more tags (Table~\ref{tab:levels}). Shaded rows mark \medaka\ (``MED'') and the performance of the best method per model is marked in bold. Appendix~\ref{app:pass3} reports \texttt{pass@3}(\%). }
\label{tab:accuracy_tags}
\renewcommand{\arraystretch}{0.95}
\footnotesize
\setlength{\tabcolsep}{4pt}
\resizebox{\textwidth}{!}{%
\begin{tabular}{llccccccc}
\toprule
\textbf{Tag} & \textbf{Method}
  & \makecell{Claude\\Sonnet 4.6} & \makecell{Gemini 3\\Flash} & \makecell{GPT\\5.4} & \makecell{GPT\\4o}  & \makecell{LLaMA 3.3\\70B} & \makecell{Gemma4\\4B} & \makecell{Gemma4\\2B} \\
\midrule
\multirow{3}{*}{linear}
  & CoT   & 95.1  & 84.8  & \textbf{91.9} & 32.1 & 31.1 & 71.8 & 32.6 \\
  & AoT+  & 96.6  & 91.2  & 90.0 & 23.5 & 20.3 & 67.4 & 28.7 \\
\rowcolor{medtint}
  \cellcolor{white} & MED  & \textbf{97.3} & \textbf{94.1}  & 90.9 & \textbf{45.1} & \textbf{50.5} & \textbf{79.9} & \textbf{37.7} \\
\midrule
\multirow{3}{*}{nonlinear}
  & CoT  & 33.3 & 16.4   & 28.9 & 9.4 & \textbf{15.1} & 14.5 & 10.1\\
  & AoT+  & 55.3 & 21.4  & \textbf{49.1} & 10.1 & 10.7 & \textbf{15.1} & 11.3 \\
\rowcolor{medtint}
  \cellcolor{white} & MED & \textbf{87.4} & \textbf{50.9} & 46.5 & \textbf{15.7} & \textbf{15.1} & 12.6 & \textbf{11.9} \\
\midrule
\multirow{3}{*}{coupling}
  & CoT   & 60.4 & 39.6  & 54.7 & 2.5 & 2.5 & 34.9 &  4.7\\
  & AoT+  & 73.6 & 49.4  & \textbf{62.9} & 2.5 & 1.6 & 28.9 & 4.4 \\
\rowcolor{medtint}
  \cellcolor{white} & MED  & \textbf{90.3} & \textbf{68.9}  & \textbf{62.9} & \textbf{10.7} & \textbf{14.2} & \textbf{38.1} & \textbf{10.4} \\
\midrule
\multirow{3}{*}{multi-goal}
  & CoT  & 50.4 & 31.1   & 43.4 & 15.8 & 14.5 & 20.2 & \textbf{14.9} \\
  & AoT+  & 65.8 & 37.3  & \textbf{56.6} & 14.5 & 10.5 & 21.9 & 11.4 \\
\rowcolor{medtint}
  \cellcolor{white} & MED  & \textbf{89.5} & \textbf{60.5}  & 54.8 & \textbf{16.7} & \textbf{18.4} & \textbf{24.6} & 9.6 \\
\midrule
\multirow{3}{*}{distractor}
  & CoT   & 73.8 & 66.7  & 72.6 & 10.7 & 25.0 & 57.1 & \textbf{34.5} \\
  & AoT+  & 88.1 & 67.9  & \textbf{83.3} & 7.1 & 15.5& 53.6 & 20.2 \\
\rowcolor{medtint}
  \cellcolor{white} & MED  & \textbf{94.0} & \textbf{77.4}  & 76.2 & \textbf{42.9} & \textbf{42.9} & \textbf{63.1} & 33.3 \\
\midrule
\multirow{3}{*}{\textbf{Total}}
  & CoT   & 77.8 & 65.6 & 74.3 & 25.7  & 26.6 & 55.7 & 26.3 \\
  & AoT+  & 85.0 & 71.6  & \textbf{78.5} & 19.8 & 17.6 & 52.7 & 23.8 \\
\rowcolor{medtint}
  \cellcolor{white} & MED  & \textbf{94.5} & \textbf{82.0}  & \textbf{78.5} & \textbf{36.9} & \textbf{40.6} & \textbf{61.0} & \textbf{30.5} \\
\bottomrule
\end{tabular}%
}
\end{table}

%% file: tables/ablation.tex
\begin{table}[t]
\centering
\caption{Ablation of \medaka\ on \texttt{Claude-Sonnet-4.6} and \texttt{Gemini-3-Flash}, 3 trials per problem.
``$-$\,Recursive Module'' drops the recursive goal decomposition module;
``$-$\,State Tracking'' drops the per-step state tracking module, which includes all numeric computation instructions.
Each cell reports \texttt{pass@1} (\%) on the full \liquidworld\, with the absolute drop (in percentage points) relative to full \medaka\ in parentheses. See Table~\ref{tab:ablation_tags} in Appendix~\ref{app:ablation_tags} for per-tag ablation results.}
\label{tab:ablation}
\setlength{\tabcolsep}{4pt}
\renewcommand{\arraystretch}{0.95}

\sisetup{table-format=2.1, retain-explicit-plus=false}
\begin{tabular}{l S@{\,}l S@{\,}l}
\toprule
& \multicolumn{2}{c}{Claude-Sonnet-4.6} & \multicolumn{2}{c}{Gemini-3-Flash} \\
\midrule
\medaka
& 94.5 &          & 82.0 &          \\
$-$\,State Tracking
& 90.5 & ($-$4.0) & 76.0 & ($-$6.0) \\
$-$\,Recursive Module
& 91.0 & ($-$3.5) & 75.1 & ($-$6.9) \\
\bottomrule
\end{tabular}
\end{table}


%% file: sections/discussion.tex
\section{Discussion}
We studied LLM planning in continuous-state domains, where actions form a continuum and do not satisfy the enumerate-and-rank assumption of existing LLM planning frameworks.
We developed \medaka, a structured prompting framework for LLM planning inspired by classical Means-Ends Analysis, and \liquidworld, a systematic testbed for LLM continuous-state planning grounded in fluid mixing, covering diverse scenarios and measuring key LLM reasoning capabilities through structural difficulty tags. 
\medaka\ can outperform existing prompting-based baselines across difficulty tags, surpass CoT's oracle-selected \texttt{pass@3} on four of seven models, and retain gains over baselines when either \medaka\ component is ablated. 
All these results corroborate that recursive goal decomposition and per-step state tracking, as encouraged by Means-Ends Analysis, both help LLMs plan in continuous-state domains.


While set in fluid mixing, \liquidworld\ is designed as a controlled testbed rather than a benchmark: it isolates parameter derivation with deterministic, fully specified transitions, so that failures can be attributed to LLM reasoning. 
Though \liquidworld already exposes systematic failures under CoT using a simple transition function, it could be extended with stochastic transitions, partial observability, or additional constraint types.
Finally, pairing \medaka\ with external solvers or verifiers is a separate direction from ours: we restrict comparisons to single-turn prompting methods because our goal is to characterize the native planning abilities of LLMs.

%% file: sections/app_all_other.tex
\section{Appendix}

\subsection{Algorithmic Description of \medaka}
The \medaka\ procedure is summarized in Algorithm~\ref{alg:med}.
\begin{algorithm}[htb]
\caption{\medaka: Means-Ends Decomposition for continuous-state LLM planning}
\label{alg:med}
\begin{algorithmic}[1]
\Require initial state $s_0$, goal $g$, LLM $\pi$, step budget $T_{\max}$, subgoal depth budget $c_{\max}$
\Ensure action sequence $a_{1:T}$ such that $g(s_T)=1$, or \textsc{Fail}
\State $\textsc{State}_0 \gets s_0$
\For{$t = 1, 2, \ldots, T_{\max}$}
  \State \textbf{[Reason]:}
  \State $D \gets \pi.\textsc{Diff}(\textsc{State}_{t-1},\, g)$
        \Comment{unsatisfied goal conditions}
  \If{$D = \emptyset$}
    \State \Return $a_{1:t-1}$
  \EndIf
  \State The LLM selects a gap $d \in D$
  \State $o \gets \pi.\textsc{ProposeOperator}(d,\, \textsc{State}_{t-1})$
  \State $c \gets 0$
  \While{$\neg\textsc{applicable}(o)$ \textbf{and} $c < c_{\max}$}
        \Comment{$\textsc{applicable}(o) \triangleq$ all preconditions met $\land$ all parameters determined}
    \State Identify blocker $\beta$ in $a$
        \Comment{unmet precondition or undetermined parameter}
    \State $g' \gets$ subgoal: ``make $\beta$ satisfied''
    \State Derive $g'$ via algebraic solving on $(\textsc{State}_{t-1},\, g)$
        \Comment{parameter values computed under per-step state tracking}
    \State Update $o$ with derived value
    \State $c \gets c + 1$
  \EndWhile
  \If{$\neg\textsc{applicable}(o)$}
    \State \Return \textsc{Fail}
  \EndIf
  \State \textbf{[Action]} $a_t \gets o$
  \State $\textsc{State}_t \gets \pi.\textsc{ComputeFullState}(\textsc{State}_{t-1},\, a_t)$
        \Comment{complete state: every vessel, not only those affected by $a_t$}
\EndFor
\State \Return \textsc{Fail} \Comment{step budget exceeded}
\end{algorithmic}
\end{algorithm}

\subsection{Pass@3 results}
\label{app:pass3}

Table~\ref{tab:pass3_tags} reports Pass@3 for every model and method, computed over the same
$n{=}3$ rollouts as Table~\ref{tab:accuracy_tags}: a problem is counted as solved if at least one rollout produces a plan accepted by the validator. 

\begin{table}[ht]
\centering
\caption{Pass@3 (\%) on \textsc{LiquidWorld} by difficulty tag. Shaded rows mark \textsc{Medaka} (``Med'').}
\label{tab:pass3_tags}
\renewcommand{\arraystretch}{1.15}
\setlength{\tabcolsep}{4pt}
\resizebox{\textwidth}{!}{%
\begin{tabular}{llccccccc}
\toprule
\textbf{Tag} & \textbf{Method}
 & \makecell{Claude\\Sonnet 4.6} & \makecell{Gemini 3\\Flash} & \makecell{GPT\\5.4} & \makecell{GPT\\4o} & \makecell{LLaMA 3.3\\70B} & \makecell{Gemma4\\4B} & \makecell{Gemma4\\2B} \\
\midrule
\multirow{3}{*}{linear}
  & CoT  & 95.6 & 86.0 & 94.9 & 41.9 & 39.0 & 79.4 & 42.6 \\
  & AoT+ & 97.8 & 91.2 & 94.9 & 29.4 & 27.9 & 77.2 & 39.0 \\
\rowcolor{medtint}
  \cellcolor{white} & MED & 98.5 & 94.1 & 94.9 & 62.5 & 68.4 & 87.5 & 54.4 \\
\midrule
\multirow{3}{*}{nonlinear}
  & CoT  & 64.2 & 18.9 & 49.1 & 11.3 & 15.1 & 17.0 & 11.3 \\
  & AoT+ & 88.7 & 22.6 & 73.6 & 11.3 & 15.1 & 15.1 & 13.2 \\
\rowcolor{medtint}
  \cellcolor{white} & MED & 96.2 & 58.5 & 75.5 & 17.0 & 15.1 & 13.2 & 13.2 \\
\midrule
\multirow{3}{*}{coupling}
  & CoT  & 76.4 & 42.5 & 67.9 & 6.6 & 5.7 & 38.7 & 9.4 \\
  & AoT+ & 91.5 & 50.0 & 81.1 & 4.7 & 3.8 & 34.0 & 9.4 \\
\rowcolor{medtint}
  \cellcolor{white} & MED & 96.2 & 72.6 & 81.1 & 18.9 & 26.4 & 44.3 & 16.0 \\
\midrule
\multirow{3}{*}{multi-goal}
  & CoT  & 72.4 & 32.9 & 60.5 & 17.1 & 17.1 & 21.1 & 15.8 \\
  & AoT+ & 89.5 & 38.2 & 77.6 & 17.1 & 13.2 & 23.7 & 13.2 \\
\rowcolor{medtint}
  \cellcolor{white} & MED & 96.1 & 65.8 & 78.9 & 25.0 & 22.4 & 27.6 & 17.1 \\
\midrule
\multirow{3}{*}{distractor}
  & CoT  & 85.7 & 67.9 & 78.6 & 32.1 & 35.7 & 64.3 & 50.0 \\
  & AoT+ & 100.0 & 67.9 & 96.4 & 21.4 & 21.4 & 60.7 & 28.6 \\
\rowcolor{medtint}
  \cellcolor{white} & MED & 100.0 & 78.6 & 89.3 & 57.1 & 57.1 & 67.9 & 50.0 \\
\midrule
\multirow{3}{*}{\textbf{Total}}
  & CoT  & 86.8 & 67.2 & 82.0 & 33.3 & 32.3 & 61.9 & 33.9 \\
  & AoT+ & 95.2 & 72.0 & 88.9 & 24.3 & 24.3 & 59.8 & 31.7 \\
\rowcolor{medtint}
  \cellcolor{white} & MED & 97.9 & 84.1 & 89.4 & 49.7 & 53.4 & 66.7 & 42.9 \\
\bottomrule
\end{tabular}%
}
\end{table}

\subsection{Per-tag ablation results}
\label{app:ablation_tags}

Table~\ref{tab:ablation_tags} breaks the ablation of Table~\ref{tab:ablation} down by difficulty tag (Table~\ref{tab:levels}). 

\begin{table}[h]
\centering
\caption{per-tag ablation results of \medaka\ (\texttt{pass@1}, \%, 3 trials per problem).
  Columns as in Table~\ref{tab:ablation}.}
\label{tab:ablation_tags}
\setlength{\tabcolsep}{5pt}
\renewcommand{\arraystretch}{1.15}
\begin{tabular}{lcccccc}
\toprule
 & \multicolumn{3}{c}{Claude-Sonnet-4.6} & \multicolumn{3}{c}{Gemini-3-Flash} \\
\cmidrule(lr){2-4}\cmidrule(lr){5-7}
\textbf{Tag} & \medaka & $-$\,State & $-$\,Recursive & \medaka & $-$\,State & $-$\,Recursive \\
\midrule
linear         & 97.3 & 93.6 & 94.1 & 94.1 & 91.4 & 93.1 \\
nonlinear      & 87.4 & 82.4 & 83.0 & 50.9 & 36.5 & 28.9 \\
coupling       & 90.3 & 88.4 & 89.3 & 68.9 & 58.2 & 56.0 \\
multi-goal     & 89.5 & 86.0 & 87.7 & 60.5 & 46.9 & 43.4 \\
distractor     & 94.0 & 90.5 & 92.9 & 77.4 & 69.0 & 67.9 \\
\midrule
\textbf{Total} & 94.5 & 90.5 & 91.0 & 82.0 & 76.0 & 75.1 \\
\bottomrule
\end{tabular}
\end{table}

\subsection{Failure Mode Example}\label{app:failure}
Table~\ref{tab:qualitative-failure-traces} shows examples for each representative failure mode. 

\input{tables/failure}

\subsection{Problem Generator: Linear Programming Solvability Formulations}
\label{app:generator}

Before any instance enters the testbed, the generator verifies solvability
via three checks (Section~\ref{sec:liquidworld}).
The first two are simple necessary conditions; the third is a linear program
that is both necessary and sufficient.

\paragraph{Volume sufficiency.}
For each liquid type $k \in \{1,\dots,K\}$, let
$a_k = \sum_{i} \mathit{vol}_i \cdot c_i^k$ be the total available volume
of liquid~$k$ across all sources, and
$d_k = \sum_{j} g_j^k$ the total demand across all goal vessels.
The check requires $a_k \ge d_k$ for every~$k$.

\paragraph{Concentration feasibility.}
For each liquid~$k$, let $\bar{c}^k = \max_i\, c_i^k$ be the maximum
fraction of~$k$ across all source vessels.
For each goal vessel~$j$ with target fraction $\hat{c}_j^k$, the check
requires $\hat{c}_j^k \le \bar{c}^k$.
This is a necessary condition because pouring preserves mixture ratios:
no sequence of pours can concentrate a liquid above its highest source
fraction.

\paragraph{Single-target LP.}
For a single goal vessel with target component volumes
$(g^1, \dots, g^K)$, the LP is:
\begin{align}
  \min_{x_1,\dots,x_n} \quad & \sum_{i=1}^{n} x_i
    \label{eq:lp-obj} \\
  \text{s.t.} \quad
  & \sum_{i=1}^{n} x_i \cdot c_i^k = g^k,
    \quad \forall\, k \in \{1,\dots,K\}
    \label{eq:lp-eq} \\
  & 0 \le x_i \le \mathit{tot}_i,
    \quad \forall\, i \in \{1,\dots,n\}
    \label{eq:lp-bounds}
\end{align}
where $x_i$ is the volume poured from source~$i$,
$\mathit{tot}_i$ is the total volume in source~$i$,
and $c_i^k$ is the fraction of liquid~$k$ in source~$i$.
The objective is a dummy minimisation; only feasibility matters.

\paragraph{Joint LP (multiple targets).}
When the goal specifies $m > 1$ target vessels that may share sources,
the decision variables become $x_{ij}$ (volume poured from source~$i$ to
target~$j$), giving $n \times m$ variables:
\begin{align}
  \min_{x_{ij}} \quad & \sum_{i,j} x_{ij}
    \label{eq:jlp-obj} \\
  \text{s.t.} \quad
  & \sum_{i=1}^{n} x_{ij} \cdot c_i^k = g_j^k,
    \quad \forall\, j \in \{1,\dots,m\},\;
    \forall\, k \in \{1,\dots,K\}
    \label{eq:jlp-eq} \\
  & \sum_{j=1}^{m} x_{ij} \le \mathit{tot}_i,
    \quad \forall\, i \in \{1,\dots,n\}
    \label{eq:jlp-cap} \\
  & x_{ij} \ge 0,
    \quad \forall\, i,j
    \label{eq:jlp-bounds}
\end{align}
Constraint~\eqref{eq:jlp-eq} requires each target to receive exactly the
right volume of each liquid component.
Constraint~\eqref{eq:jlp-cap} ensures that the total volume drawn from each
source across all targets does not exceed its available volume.
An instance is admitted to the testbed only if the LP returns a feasible
solution.
\newpage
\subsection{Token Usage Details}\label{app:token}
We report per-trial token usage for all methods and models for \liquidworld\ in Table~\ref{tab:tokens}. 
\input{tables/tokens}

%% file: tables/failure.tex
\begin{table}[t]
\centering
\small
\caption{Representative failed traces illustrating three qualitative failure modes.}
\label{tab:qualitative-failure-traces}
\begin{tabular}{p{0.94\linewidth}}
\toprule
\textbf{Failure mode 1: Trial-and-error repair after volume saturation} \\
\midrule

\textbf{Example}
Initial state: 
$V_1$: $10$ L $L_2$; 
$V_2$: $10$ L $(40\%L_1,60\%L_3)$; 
$V_3$: $10$ L $L_3$; 
$V_4$: empty. \\
Goal: $V_4=2$ L $(13.43\%L_1,30.52\%L_2,56.05\%L_3)$. \\

\texttt{pour 1.0L V2 -> V4}
\quad $\Rightarrow$ \quad
$V_4: 1$ L, $(40,0,60)\%$
\quad  \\

\texttt{pour 0.5L V1 -> V4}
\quad $\Rightarrow$ \quad
$V_4: 1.5$ L, $(26.7,33.3,40)\%$
\quad \\

\texttt{pour 0.5L V3 -> V4}
\quad $\Rightarrow$ \quad
$V_4: 2$ L, $(20,25,55)\%$
\quad \textit{volume reached, composition wrong} \\

\texttt{pour 0.1L V2 -> V4}
\quad $\Rightarrow$ \quad
$V_4: 2.1$ L, $(21.0,23.8,55.2)\%$
\quad \textit{repair after saturation} \\

\texttt{pour 0.1L V1 -> V4}
\quad $\Rightarrow$ \quad
$V_4: 2.2$ L, $(20,27.3,52.7)\%$
\quad \textit{continued local repair} \\

\texttt{pour 0.2L V4 -> V1}
\quad $\Rightarrow$ \quad
$V_4: 2$ L, $(20,27.3,52.7)\%$
\quad \textit{attempted undo by pouring mixture back} \\

\midrule
\textbf{Failure mode 2: State hallucination from source depletion} \\
\midrule

\textbf{Example}
Initial state: 100 source vessels, each $1$ L pure liquid; 
$V_8$ contain $L_4$. \\
Goal: $V_{34}$= 2L (50\%L4, 50\%L12), $V_{51}$ = 2L (50\%L15, 50\%L13) $\dots$; 
$V_{82}$ = 2L (50\%L20, 50\%L4). \\

\texttt{pour 1L V12 -> V34}
\quad $\Rightarrow$ \quad
$V_{12}$ empty \\

\texttt{pour 1L V8 -> V34}
\quad $\Rightarrow$ \quad
$V_8$ empty
\quad \textit{$L_4$ source depleted} \\

\texttt{pour 1L V50 -> V51}
\quad $\Rightarrow$ \quad
$V_{50}$ empty \\

\texttt{pour 1L V15 -> V51}
\quad $\Rightarrow$ \quad
$V_{15}$ empty \\

\texttt{pour 1L V13 -> V82}
\quad $\Rightarrow$ \quad
$V_{13}$ empty \\

\texttt{pour 1L V8 -> V82}
\quad $\Rightarrow$ \quad
\textit{invalid action: reuses already-empty $V_8$} \\

\midrule
\textbf{Failure mode 3: Local correctness without global mass balance} \\
\midrule
\textbf{Example}
Initial state: 
$V_1$: $10$ L $(40\%L_1,60\%L_2)$; 
$V_2$: $10$ L $(70\%L_1,30\%L_3)$; 
$V_3$: $10$ L $L_3$; 
$V_4$: empty. 
Goal: $V_4=4$ L with $(36.71\%L_1,26.67\%L_2,36.62\%L_3)$. \\
\texttt{pour 2.667L V1 -> V4}
\quad $\Rightarrow$ \quad
$V_4: 2.667$ L, $(40,60,0)\%$
\quad \\
\texttt{pour 0.573L V2 -> V4}
\quad $\Rightarrow$ \quad
$V_4: 3.24$ L, $(45.3,49.4,5.3)\%$
\quad  \\
\texttt{pour 0.76L V3 -> V4}
\quad $\Rightarrow$ \quad
$V_4: 4$ L, $(36.7,40.0,23.3)\%$
\quad \textit{volume reached, concentration not} \\
\bottomrule
\end{tabular}
\end{table}

%% file: tables/tokens.tex
\label{sec:appendix-tokens}

\begin{table}[h]
\centering
\caption{\textbf{Per-trial token usage on \textsc{LiquidWorld} }.
Each cell reports the median (and Q1--Q3 interquartile range) across all problems and trials.}
\label{tab:tokens}
\small
\setlength{\tabcolsep}{4pt}
\begin{tabular}{ll lll}
\toprule
Model & Method & Input & Output & Total \\
\midrule
\multirow{3}{*}{Claude-4.6} & CoT & 962 {\scriptsize (876--1425)} & 1231 {\scriptsize (1019--1392)} & 2185 {\scriptsize (1898--2931)} \\
 & AoT+ & 1463 {\scriptsize (1357--1873)} & 1726 {\scriptsize (1342--15891)} & 3134 {\scriptsize (2734--17580)} \\
 & MED & 2381 {\scriptsize (2269--2831)} & 4498 {\scriptsize (2432--17725)} & 6857 {\scriptsize (4679--20482)} \\
\midrule
\multirow{3}{*}{Gemini3-Flash} & CoT & 931 {\scriptsize (839--1448)} & 1028 {\scriptsize (892--1393)} & 1989 {\scriptsize (1731--2855)} \\
 & AoT+ & 1308 {\scriptsize (1181--1874)} & 1225 {\scriptsize (1032--1776)} & 2543 {\scriptsize (2250--3746)} \\
 & MED & 2206 {\scriptsize (2079--2764)} & 1888 {\scriptsize (1555--2954)} & 4053 {\scriptsize (3665--5685)} \\
\midrule
\multirow{3}{*}{GPT-5.4} & CoT & 872 {\scriptsize (794--1313)} & 966 {\scriptsize (626--1516)} & 1859 {\scriptsize (1446--2777)} \\
 & AoT+ & 1206 {\scriptsize (1104--1751)} & 1415 {\scriptsize (998--3438)} & 2640 {\scriptsize (2052--5152)} \\
 & MED & 2083 {\scriptsize (1981--2674)} & 4981 {\scriptsize (3209--8815)} & 7013 {\scriptsize (5186--10987)} \\
\midrule
\multirow{3}{*}{GPT-4o} & CoT & 873 {\scriptsize (795--1314)} & 915 {\scriptsize (771--1108)} & 1849 {\scriptsize (1617--2474)} \\
 & AoT+ & 1247 {\scriptsize (1118--2053)} & 1121 {\scriptsize (813--1640)} & 2551 {\scriptsize (1945--3536)} \\
 & MED & 2084 {\scriptsize (1982--2605)} & 2116 {\scriptsize (1582--3277)} & 4402 {\scriptsize (3546--6069)} \\
\midrule
\multirow{3}{*}{Llama-3.3-70B} & CoT & 926 {\scriptsize (827--1630)} & 1151 {\scriptsize (879--1404)} & 2189 {\scriptsize (1847--2854)} \\
 & AoT+ & 1251 {\scriptsize (1139--1888)} & 1828 {\scriptsize (1396--2949)} & 3507 {\scriptsize (2754--4734)} \\
 & MED & 2110 {\scriptsize (2008--2714)} & 2029 {\scriptsize (1527--2707)} & 4290 {\scriptsize (3843--5116)} \\
\midrule
\multirow{3}{*}{Gemma4-4B} & CoT & 957 {\scriptsize (865--1464)} & 2735 {\scriptsize (2153--3700)} & 4121 {\scriptsize (3126--5098)} \\
 & AoT+ & 1324 {\scriptsize (1196--1900)} & 3672 {\scriptsize (2793--5127)} & 5466 {\scriptsize (4145--6872)} \\
 & MED & 2221 {\scriptsize (2105--2790)} & 5473 {\scriptsize (3944--7690)} & 7839 {\scriptsize (6284--10539)} \\
\midrule
\multirow{3}{*}{Gemma4-2B} & CoT & 956 {\scriptsize(863--1462)}   & 3355 {\scriptsize(2486--3895)} & 4512 {\scriptsize(3617--5313)} \\
& AoT+ & 1322 {\scriptsize(1195--1888)} & 3805 {\scriptsize(3089--4534)} & 5488 {\scriptsize(4507--6229)} \\
& MED  & 2220 {\scriptsize(2093--2778)} & 5739 {\scriptsize(4375--7590)} & 8219 {\scriptsize(6775--10188)} \\
\bottomrule
\end{tabular}
\end{table}

%% file: sections/app_prompts.tex
\clearpage
\section{Implementation Details}
We report the complete system and user prompts for all methods tested. These prompts are identical across all models.
\subsection{Complete LLM Prompt}
We provide the system and user prompts for AoT+, \medaka, and CoT. All proprietary models (GPT-4o, GPT-5.4, Claude Sonnet 4.6, Gemini 3 Flash) were accessed via their providers' official APIs under standard terms of service. The open-weight models LLaMA-3.3-70B (Llama 3.3 Community License) and Gemma4-2B/4B (Gemma Terms of Use) were accessed via Featherless AI under standard terms of service. We use temperature=0 for the rollouts. 
\label{app:prompt}

\newtcblisting{systemprompt}[1]{
  breakable,
  listing only,
  title=#1,
  colback=blue!8,
  colframe=blue!60,
  colbacktitle=blue!60,
  coltitle=white,
  arc=4pt,
  fonttitle=\bfseries\small,
  listing options={
    basicstyle=\ttfamily\small,
    breaklines=true,
    breakatwhitespace=false,
    columns=fullflexible,
    keepspaces=true,
  },
}

\newtcblisting{userprompt}[1]{
  breakable,
  listing only,
  title=#1,
  colback=orange!10,
  colframe=orange!70,
  colbacktitle=orange!70,
  coltitle=white,
  arc=4pt,
  fonttitle=\bfseries\small,
  listing options={
    basicstyle=\ttfamily\small,
    breaklines=true,
    breakatwhitespace=false,
    columns=fullflexible,
    keepspaces=true,
  },
}

\begin{systemprompt}{System prompt}
You are a planning agent.

## Domain Description:

# LiquidWorld Planning Domain

## Description:
You have vessels with fixed capacities. Vessels hold homogeneous mixtures of liquids (L1, L2, L3, ...).
All liquids have the same density and blend when mixed. You pour a specified volume from one vessel to another.

## Rules:
1. Pour only from non-empty vessels.
2. Specify volume in L to pour (e.g., 0.5 L).
3. Poured liquid preserves the mixture ratio of the source.
When pouring from a mixture, ALL liquid types transfer in proportion to their concentration.
4. Destination must have enough remaining capacity.
5. Any non-empty vessel can serve as a source, including a goal vessel that has been partially filled.
A goal vessel only needs to match its target state at the END of the plan; intermediate contents are unconstrained.

## State and Goal Format:
- Pure liquid: Vi (capacity x L): y L Ln
- Mixture: Vi (capacity x L): y L, z% L1, w% L2
- Empty: Vi (capacity x L): empty
- Goal: Vi: x L, y% L1, z% L2

## Action Format:
- Pour x L from Vi to Vj
\end{systemprompt}

\begin{userprompt}{AoT+ user prompt}
# Algorithm of Thoughts Plus (AoT+) Planning System

## Planning Task:
Given an initial state and a goal state, create a plan to transform the initial state into the goal state.
Use the Algorithm of Thoughts Plus approach to solve this planning problem.

### Key Instructions:
1. Think step by step to explore possible actions
2. Periodically track and restate the current state (using STATE notation)
3. Explore different paths when needed
4. Backtrack if a path doesn't seem promising
5. Build towards the goal state systematically
6. For each action you take, start the line with "ACTION:" followed by the action description
7. After each action, show the new state using "STATE x.y:" notation
8. Conclude with a "Plan summary:" section that lists all the actions in numbered format

## Examples:
Initial state: V1 (capacity 1L): 1 L L1; V2 (capacity 1L): 1 L L2; V3 (capacity 2L): empty
Goal state: V1: 0.5 L L1; V2: 0.5 L L2; V3: 1 L, 50% L1, 50% L2

STATE 1: Pour 4 L from V1 to V3
Let's try a different approach.

STATE 2: Pour 0.5 L from V1 to V3
STATE 2.1: Pour 1 L from V2 to V3

This approach isn't working well. Let's backtrack from STATE 2
STATE 2.2: Pour 0.5 L from V2 to V3

Final check: V1: 0.5 L L1V2: 0.5 L L2V3: 1 L, 50% L1, 50% L2 has been achieved.
...
Plan summary:
1. Pour 0.5 L from V1 to V3
2. Pour 0.5 L from V2 to V3

---

## Problem to Solve:

Initial state: V1 (capacity 999L): 10 L L1; V2 (capacity 999L): 10 L L2; V3 (capacity 999L): 10 L L3; V4 (capacity 999L): empty
Goal state: V4: 5 L, 40% L1, 40% L2, 20% L3

STATE 0: V1 (capacity 999L): 10 L L1; V2 (capacity 999L): 10 L L2; V3 (capacity 999L): 10 L L3; V4 (capacity 999L): empty

Solve this step by step, tracking the state periodically. Remember to:
1. Prefix each action with "ACTION:"
2. Update the state after significant changes using "STATE x.y:" notation
3. End with a "Plan summary:" section

Start your solution:
\end{userprompt}

\begin{userprompt}{\medaka\ user prompt}
# MEDAKA Planning
## Planning Task:
Given an initial state and a goal state, create a plan to transform the initial state into the goal state.
Use the MEA planning format below.

### MEA Planning Format 

Think step by step to explore possible actions.
You MUST follow this format strictly for every planning step.

### Key Instructions:
1. Compare the current state with the goal state and list all differences (Diff)
2. Pick one difference and search for an operator that reduces it
3. Judge whether the operator is directly solvable
(solvable = all preconditions met AND all parameters determined)
4. If not solvable, set "make the operator solvable" as a subgoal
and repeat from step 2 for that subgoal until a solvable operator is found
5. Execute the solvable operator, update the state, and return to step 1 for remaining differences
6. If the result deviates from expectation, diagnose the cause and adjust.

Each step:
[REASON]
- Diff: compare current state (from STATE k-1) with goal and list all differences (Diff)
- Operator: pick one action to reduce a gap; judge whether it is directly solvable
- Subgoal: what blocks the operator? State the blocker.
First, list each operand's per-unit contribution to the result.
Then set up equations, solve for parameters, and confirm all parts
are fully derived and the action moves closer to the goal.
Repeat (Subgoal -> Operator) until a solvable operator is found.
[ACTION] State the solvable operator with its computed parameter
STATE k: Write out the FULL new state (k is the step number).
You may show brief arithmetic before the STATE line when computing
new compositions for mixtures.

End your solution with a "Plan summary:" section listing only the actions from your successful path.

## Example:
Initial state: V1 (capacity 1L): 1 L L1; V2 (capacity 1L): 1 L L2; V3 (capacity 2L): empty
Goal state: V3: 1 L, 50% L1, 50% L2

STATE 0: V1 (capacity 1L): 1 L L1; V2 (capacity 1L): 1 L L2; V3 (capacity 2L): empty

STEP 1:
[REASON]
- Diff: V3 empty, goal 0.5L L1 + 0.5L L2 | V1 excess 0.5L | V2 excess 0.5L
- Operator: Pour to V3 -- not solvable (V3 needs L1 + L2; source unknown)
- Subgoal: determine sources. Sources with L1: V1 only. Sources with L2: V2 only.
No shared components across sources -> solve one at a time. Start with L1.
-> Operator: Pour from V1 to V3 -- not solvable (amount unknown)
-> Subgoal: V3 needs (1L * 50% = 0.5L) L1; V1 is pure L1 (concentration = 100%) -> pour amount = 0.5 / 1.0 = 0.5L
-> Operator: Pour 0.5L from V1 to V3 -- solvable (V1 has 1L >= 0.5L, V3 cap 2L >= 0.5L)
[ACTION] Pour 0.5 L from V1 to V3
STATE 1:
V1: 1 - 0.5 = 0.5L L1
V3 receives 0.5L pure L1: L1 = 0.5*100% = 0.5L; total = 0.5L
V1: 0.5L L1 | V2: 1L L2 | V3: 0.5L L1

STEP 2:
[REASON]
- Diff: V3 has 0.5L L1 + 0L L2, goal 0.5L L1 + 0.5L L2 -> remaining: 0.5L L2
- Operator: Pour to V3 -- not solvable (amount unknown)
- Subgoal: Sources with L2: V2 only (pure L2). Amount unknown.
V3 needs (1L * 50% = 0.5L) L2; V2 is pure L2 (concentration = 100%) -> pour amount = 0.5 / 1.0 = 0.5L
-> Operator: Pour 0.5L from V2 to V3 -- solvable (V2 has 1L >= 0.5L, V3 cap 2L >= 0.5L)
[ACTION] Pour 0.5 L from V2 to V3
STATE 2:
V2: 1 - 0.5 = 0.5L L2
V3 receives 0.5L pure L2: L2 = 0.5*100% = 0.5L; total = 0.5 + 0.5 = 1L, L1 = 0.5/1 = 50%, L2 = 0.5/1 = 50%
V1: 0.5L L1 | V2: 0.5L L2 | V3: 1L 50%L1 50%L2

Plan summary:
1. Pour 0.5 L from V1 to V3
2. Pour 0.5 L from V2 to V3

---

## Problem to Solve:

Initial state: V1 (capacity 999L): 10 L L1; V2 (capacity 999L): 10 L L2; V3 (capacity 999L): 10 L L3; V4 (capacity 999L): empty
Goal state: V4: 5 L, 40% L1, 40% L2, 20% L3

STATE 0: V1 (capacity 999L): 10 L L1; V2 (capacity 999L): 10 L L2; V3 (capacity 999L): 10 L L3; V4 (capacity 999L): empty

Solve this step by step using the MEA planning format above. Remember:
1. For each step, use the MEA loop inside [REASON], then [ACTION] -> STATE k:
2. End with a "Plan summary:" section

Start your solution:
\end{userprompt}

\begin{userprompt}{CoT user prompt}
# Vanilla Chain-of-Thought Planning

## Planning Task:
Given an initial state and a goal state, create a plan to reach the goal.

Think step by step. End with a "Plan summary:" section listing only your final actions.

## Example:
Initial state: V1 (capacity 1L): 1 L L1; V2 (capacity 1L): 1 L L2; V3 (capacity 2L): empty
Goal state: V1: 0.5 L L1; V2: 0.5 L L2; V3: 1 L, 50% L1, 50% L2

V3 needs 1L at 50/50, so it needs 0.5L of L1 and 0.5L of L2. V1 is pure L1 with 1L, and the goal wants V1 to keep 0.5L, so I pour 0.5L from V1 to V3. Now V3 has 0.5L L1. Similarly V2 is pure L2 and should keep 0.5L, so I pour 0.5L from V2 to V3. That gives V3 = 1L at 50/50, and both V1 and V2 keep 0.5L. Done.

Plan summary:
1. Pour 0.5 L from V1 to V3
2. Pour 0.5 L from V2 to V3

---

## Problem to Solve:
Initial state: V1 (capacity 999L): 10 L L1; V2 (capacity 999L): 10 L L2; V3 (capacity 999L): 10 L L3; V4 (capacity 999L): empty
Goal state: V4: 5 L, 40% L1, 40% L2, 20% L3

Think step by step. Finish with:
Plan summary:
1. ...
\end{userprompt}

%% file: references.bib
@article{wei2022chain,
  title={Chain-of-thought prompting elicits reasoning in large language models},
  author={Wei, Jason and Wang, Xuezhi and Schuurmans, Dale and Bosma, Maarten and Xia, Fei and Chi, Ed and Le, Quoc V and Zhou, Denny and others},
  journal={Advances in neural information processing systems},
  volume={35},
  pages={24824--24837},
  year={2022}
}

@article{yao2023tree,
  title={Tree of thoughts: Deliberate problem solving with large language models},
  author={Yao, Shunyu and Yu, Dian and Zhao, Jeffrey and Shafran, Izhak and Griffiths, Tom and Cao, Yuan and Narasimhan, Karthik},
  journal={Advances in neural information processing systems},
  volume={36},
  pages={11809--11822},
  year={2023}
}

@inproceedings{hao2023reasoning,
  title={Reasoning with language model is planning with world model},
  author={Hao, Shibo and Gu, Yi and Ma, Haodi and Hong, Joshua and Wang, Zhen and Wang, Daisy and Hu, Zhiting},
  booktitle={Proceedings of the 2023 Conference on Empirical Methods in Natural Language Processing},
  pages={8154--8173},
  year={2023}
}

@inproceedings{sel2024aot,
  title={Algorithm of thoughts: enhancing exploration of ideas in large language models},
  author={Sel, Bilgehan and Al-Tawaha, Ahmad and Khattar, Vanshaj and Jia, Ruoxi and Jin, Ming},
  booktitle={Proceedings of the 41st International Conference on Machine Learning},
  pages={44136--44189},
  year={2024}
}

@inproceedings{sel2025llms,
  title     = {{LLM}s Can Plan Only If We Tell Them},
  author    = {Sel, Bilgehan and Jia, Ruoxi and Jin, Ming},
  booktitle = {Proceedings of the International Conference on Learning Representations (ICLR)},
  year      = {2025},
  url       = {https://arxiv.org/abs/2501.13545}
}

@inproceedings{wang2024llm3,
  title={Llmˆ3: Large language model-based task and motion planning with motion failure reasoning},
  author={Wang, Shu and Han, Muzhi and Jiao, Ziyuan and Zhang, Zeyu and Wu, Ying Nian and Zhu, Song-Chun and Liu, Hangxin},
  booktitle={2024 IEEE/RSJ international conference on intelligent robots and systems (IROS)},
  pages={12086--12092},
  year={2024},
  organization={IEEE}
}

@inproceedings{curtis2024proc3s,
  title     = {Trust the {PR}o{C3S}: Solving Long-Horizon Robotics Problems with {LLM}s and Constraint Satisfaction},
  author    = {Curtis, Aidan and Kumar, Nishanth and Cao, Jing and Lozano-P{\'e}rez, Tom{\'a}s and Kaelbling, Leslie Pack},
  booktitle={Conference on Robot Learning},
  pages={1362--1383},
  year={2025},
  organization={PMLR}
}

@article{newell1963gps,
  title={{GPS}, a program that simulates human thought},
  author={Newell, Allen and Simon, Herbert A},
  booktitle={Computation \& Intelligence: collected readings},
  pages={415--428},
  year={1995}
}

@article{fikes1971strips,
  title={{STRIPS}: A new approach to the application of theorem proving to problem solving},
  author={Fikes, Richard E and Nilsson, Nils J},
  journal={Artificial intelligence},
  volume={2},
  number={3-4},
  pages={189--208},
  year={1971},
  publisher={Elsevier}
}

@techreport{mcdermott1998pddl,
  title={{PDDL}—the planning domain definition language},
  author={Aeronautiques, Constructions and Howe, Adele and Knoblock, Craig and McDermott, ISI Drew and Ram, Ashwin and Veloso, Manuela and Weld, Daniel and Sri, David Wilkins and Barrett, Anthony and Christianson, Dave and others},
  journal={Technical Report, Tech. Rep.},
  year={1998}
}

@article{fox2003pddl21,
  title={{PDDL2. 1}: An extension to {PDDL} for expressing temporal planning domains},
  author={Fox, Maria and Long, Derek},
  journal={Journal of artificial intelligence research},
  volume={20},
  pages={61--124},
  year={2003}
}

@article{hoffmann2003metricff,
    title={The {Metric-FF} Planning System: Translating``Ignoring Delete Lists''to Numeric State Variables},
  author={Hoffmann, J{\"o}rg},
  journal={Journal of artificial intelligence research},
  volume={20},
  pages={291--341},
  year={2003}
}

@inproceedings{erol1994htn,
  title     = {{HTN} Planning: Complexity and Expressivity},
  author    = {Erol, Kutluhan and Hendler, James and Nau, Dana S.},
  booktitle = {Proceedings of the Twelfth National Conference on Artificial Intelligence (AAAI)},
  pages     = {1123--1128},
  year      = {1994}
}

@book{ghallab2004automated,
  title={Automated Planning: theory and practice},
  author={Ghallab, Malik and Nau, Dana and Traverso, Paolo},
  year={2004},
  publisher={Elsevier}
}

@inproceedings{valmeekam2023planbench,
  title={Planbench: An extensible benchmark for evaluating large language models on planning and reasoning about change},
  author={Valmeekam, Karthik and Marquez, Matthew and Olmo, Alberto and Sreedharan, Sarath and Kambhampati, Subbarao},
  journal={Advances in Neural Information Processing Systems},
  volume={36},
  pages={38975--38987},
  year={2023}
}

@inproceedings{kambhampati2024llmmodulo,
  title={Position: {LLMs} can’t plan, but can help planning in {LLM}-modulo frameworks},
  author={Kambhampati, Subbarao and Valmeekam, Karthik and Guan, Lin and Verma, Mudit and Stechly, Kaya and Bhambri, Siddhant and Saldyt, Lucas Paul and Murthy, Anil B},
  booktitle={Forty-first International Conference on Machine Learning},
  year={2024}
}

@article{boiko2023autonomous,
  title={Autonomous chemical research with large language models},
  author={Boiko, Daniil A and MacKnight, Robert and Kline, Ben and Gomes, Gabe},
  journal={Nature},
  volume={624},
  number={7992},
  pages={570--578},
  year={2023},
  publisher={Nature Publishing Group UK London}
}

@article{m2024augmenting,
  title={Augmenting large language models with chemistry tools},
  author={M. Bran, Andres and Cox, Sam and Schilter, Oliver and Baldassari, Carlo and White, Andrew D and Schwaller, Philippe},
  journal={Nature machine intelligence},
  volume={6},
  number={5},
  pages={525--535},
  year={2024},
  publisher={Nature Publishing Group UK London}
}

@article{singh2025openai,
  title={Openai gpt-5 system card},
  author={Singh, Aaditya and Fry, Adam and Perelman, Adam and Tart, Adam and Ganesh, Adi and El-Kishky, Ahmed and McLaughlin, Aidan and Low, Aiden and Ostrow, AJ and Ananthram, Akhila and others},
  journal={arXiv preprint arXiv:2601.03267},
  year={2025}
}

@article{hurst2024gpt,
  title={Gpt-4o system card},
  author={Hurst, Aaron and Lerer, Adam and Goucher, Adam P and Perelman, Adam and Ramesh, Aditya and Clark, Aidan and Ostrow, AJ and Welihinda, Akila and Hayes, Alan and Radford, Alec and others},
  journal={arXiv preprint arXiv:2410.21276},
  year={2024}
}

@misc{anthropic2026sonnet46,
  author       = {{Anthropic}},
  title        = {Introducing {Claude onnet 4.6}},
  year         = {2026},
  month        = feb,
  day          = {17},
  howpublished = {\url{https://www.anthropic.com/news/claude-sonnet-4-6}},
  note         = {Large language model. Model ID: claude-sonnet-4-6. Accessed 2026-09-20},
  url          = {https://www.anthropic.com/news/claude-sonnet-4-6}
}

@misc{google2025gemini3flash,
  author       = {{Google}},
  title        = {{Gemini 3 Flash}: frontier intelligence built for speed},
  year         = {2025},
  month        = dec,
  day          = {17},
  howpublished = {\url{https://blog.google/products/gemini/gemini-3-flash}},
  note         = {Large language model. Accessed 2026-09-20},
  url          = {https://blog.google/products/gemini/gemini-3-flash}
}

@article{grattafiori2024llama,
  title={The {Llama} 3 herd of models},
  author={Grattafiori, Aaron and Dubey, Abhimanyu and Jauhri, Abhinav and Pandey, Abhinav and Kadian, Abhishek and Al-Dahle, Ahmad and Letman, Aiesha and Mathur, Akhil and Schelten, Alan and Vaughan, Alex and others},
  journal={arXiv preprint arXiv:2407.21783},
  year={2024}
}

@article{team2026gemma,
  title={Gemma 4 technical report},
  author={Team, Gemma and Abd, Sherif El and Aggarwal, Vaibhav and Algayres, Robin and Andreev, Alek and Bachem, Olivier and Ballantyne, Ian and Brick, Cormac and C{\u{a}}rbune, Victor and Casbon, Michelle and others},
  journal={arXiv preprint arXiv:2607.02770},
  year={2026}
}

@article{haverly1978studies,
  title={Studies of the behavior of recursion for the pooling problem},
  author={Haverly, Co A},
  journal={Acm sigmap bulletin},
  number={25},
  pages={19--28},
  year={1978},
  publisher={ACM New York, NY, USA}
}

@inproceedings{zellers2019hellaswag,
  title={Hellaswag: Can a machine really finish your sentence?},
  author={Zellers, Rowan and Holtzman, Ari and Bisk, Yonatan and Farhadi, Ali and Choi, Yejin},
  booktitle={Proceedings of the 57th annual meeting of the association for computational linguistics},
  pages={4791--4800},
  year={2019}
}

@article{chen2021evaluating,
  title={Evaluating large language models trained on code},
  author={Chen, Mark and Tworek, Jerry and Jun, Heewoo and Yuan, Qiming and Pinto, Henrique Ponde De Oliveira and Kaplan, Jared and Edwards, Harri and Burda, Yuri and Joseph, Nicholas and Brockman, Greg and others},
  journal={arXiv preprint arXiv:2107.03374},
  year={2021}
}

@article{alfaki2013strong,
  title={Strong formulations for the pooling problem},
  author={Alfaki, Mohammed and Haugland, Dag},
  journal={Journal of Global Optimization},
  volume={56},
  number={3},
  pages={897--916},
  year={2013}
}

@article{haugland2016complexity,
  title={The computational complexity of the pooling problem},
  author={Haugland, Dag},
  journal={Journal of Global Optimization},
  volume={64},
  number={2},
  pages={199--215},
  year={2016}
}

@inproceedings{ferdosi2023reinforcement,
  author    = {Ferdosi, Mohsen and Ge, Yuejun and Kingsford, Carl},
  title     = {Reinforcement Learning for Robotic Liquid Handler Planning},
  booktitle = {23rd International Workshop on Algorithms in Bioinformatics ({WABI} 2023)},
  series    = {Leibniz International Proceedings in Informatics (LIPIcs)},
  volume    = {273},
  pages     = {23:1--23:16},
  year      = {2023},
  doi       = {10.4230/LIPIcs.WABI.2023.23}
}
